\documentclass[lettersize,journal]{IEEEtran}
\usepackage{amsmath,amsfonts}
\usepackage{algorithmic}
\usepackage{array}
\usepackage[caption=false,font=normalsize,labelfont=sf,textfont=sf]{subfig}
\usepackage{textcomp}
\usepackage{stfloats}
\usepackage{url}
\usepackage{verbatim}
\usepackage{graphicx}
\def\BibTeX{{\rm B\kern-.05em{\sc i\kern-.025em b}\kern-.08em
    T\kern-.1667em\lower.7ex\hbox{E}\kern-.125emX}}
\usepackage{balance}

\usepackage{graphicx} 
\usepackage{amsmath}  
\usepackage{bbding}   
\usepackage{multirow} 
\usepackage[table]{xcolor} 

\begin{document}
\title{Revisiting Radar Camera Alignment by Contrastive Learning for 3D Object Detection}
\author{Linhua Kong, Dongxia Chang, Lian Liu, Zisen Kong, Pengyuan Li and Yao Zhao, ~\IEEEmembership{Fellow~IEEE}
\thanks{Manuscript created October, 2020; This work was developed by the IEEE Publication Technology Department. This work is distributed under the \LaTeX \ Project Public License (LPPL) ( http://www.latex-project.org/ ) version 1.3. A copy of the LPPL, version 1.3, is included in the base \LaTeX \ documentation of all distributions of \LaTeX \ released 2003/12/01 or later. The opinions expressed here are entirely that of the author. No warranty is expressed or implied. User assumes all risk.}}

\markboth{Journal of \LaTeX\ Class Files,~Vol.~18, No.~9, September~2020}%
{How to Use the IEEEtran \LaTeX \ Templates}

\maketitle

\begin{abstract}
Recently, 3D object detection algorithms based on radar and camera fusion have shown excellent performance, setting the stage for their application in autonomous driving perception tasks. Existing methods have focused on dealing with feature misalignment caused by the domain gap between radar and camera. However, existing methods either neglect inter-modal features interaction during alignment or fail to effectively align features at the same spatial location across modalities. To alleviate the above problems, we propose a new alignment model called Radar Camera Alignment (RCAlign). Specifically, we design a Dual-Route Alignment (DRA) module based on contrastive learning to align and fuse the features between radar and camera. Moreover, considering the sparsity of radar BEV features, a Radar Feature Enhancement (RFE) module is proposed to improve the densification of radar BEV features with the knowledge distillation loss. Experiments show RCAlign achieves a new state-of-the-art on the public nuScenes benchmark in radar camera fusion for 3D Object Detection. Furthermore, the RCAlign achieves a significant performance gain (4.3\% NDS and 8.4\% mAP) in real-time 3D detection compared to the latest state-of-the-art method (RCBEVDet). 
\end{abstract}


\section{Introduction} \label{Introduction}
\IEEEPARstart{A}{utonomous} driving systems are designed to enable vehicles to perform driving tasks without human intervention. Therefore, it is crucial to accurately perceive the environment around the vehicle. 3D object detection is an important research area in autonomous driving perception as it allows for accurate localization and classification of objects in the surrounding environment. Recently, mutil-view 3D object detection \cite{huang2021bevdet, liu2022petr, wang2023exploring, li2024bevnext, shu20233dppe, chen2024graph} has made significant strides in performance. Currently, cameras are widely used in practice as the main acquisition device for 3D object detection. However, cameras perform poorly in harsh environments and do not model the depth of objects well \cite{wolters2024unleashing}. Radar is used as an auxiliary sensor because of its suitability for addressing the above mentioned problems and its low cost. 

By combining the rich semantic information provided by cameras with the accurate depth information captured by the radar can lead to more robust and reliable results for 3D object detection. However, due to the domain gap between modalities, how to accurately align the same object acquired by radar and camera sensors in the fusion process is very crucial. To better align the data of these two modalities, researchers have proposed a number of alignment methods, which can be broadly classified into into two groups, dense bird-eye-view (BEV) alignment methods \cite{lin2024rcbevdet} and sparse BEV alignment methods \cite{chen2023futr3d, yu2023sparsefusion3d, kim2023crn}. 

\begin{figure*}[t]
  \centering
  \includegraphics[width=0.95\textwidth]{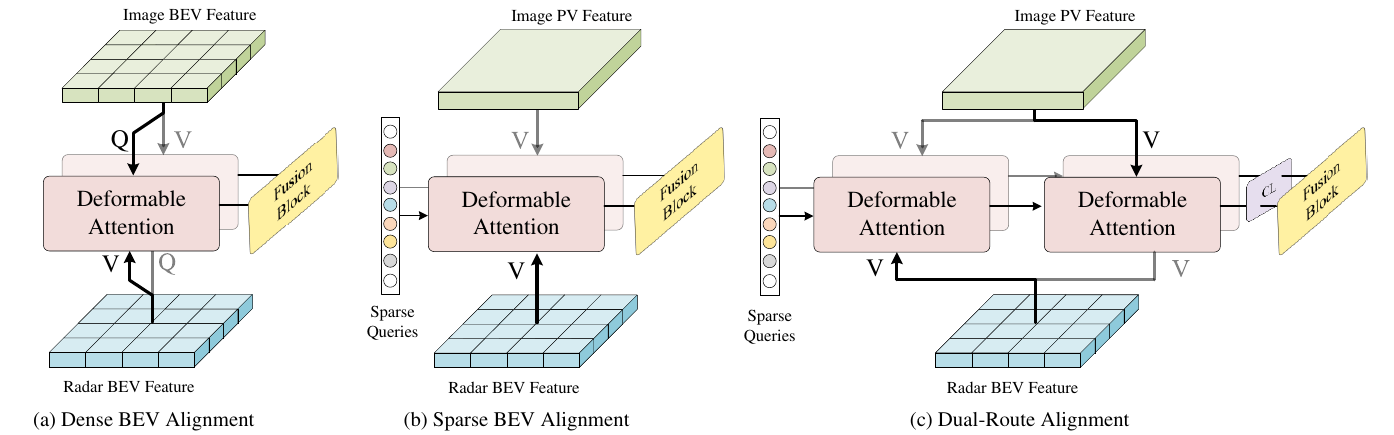}
  \caption{Different alignment methods: Dense BEV Alignment (a), Sparse BEV Alignment (b) and Dual-Route Alignment (ours c). Q, V, BEV, PV and CL indicate Query, Value, Bird-Eye-View, Perception-View and Contrastive Loss separately. The semi-transparent part and fully solid part form their own systems.}
  \label{alignment}
\end{figure*}

The Dense BEV alignment methods \cite{lin2024rcbevdet} use deformable attention to update features of one modality from another. As shown in Fig.~\ref{alignment} (a), those methods use image BEV features (Query) to find features that represent the same object in radar BEV features (Value) by deformable attention \cite{zhu2020deformable}, which is then used to update the corresponding image BEV features. The above process is also applied to radar BEV features (Query). The deformable attention is effective at correlating features representing the same object of different modalities, but it can not guarantee that the updated features of two modalities precisely represent the same object at corresponding locations. It is sub-optimal when using the fusion block to fuse them. Different from these approaches, the sparse BEV alignment methods \cite{chen2023futr3d, yu2023sparsefusion3d, kim2023crn} uses deformable attention to extract features from the two modal data separately, which are then used to update the sparse queries. Specifically, as shown in Fig.~\ref{alignment} (b), these methods use the sparse queries to model the objects in the environment, the deformable attention is then employed to find related features in image features and radar BEV features to update the sparse queries respectively. Using sparse queries as an intermediate pivot can be advantageous for aligning features representing the same object in radar and images. However, they do not consider features interaction between modalities. In summary, existing alignment methods either do not consider inter-modal features interaction during alignment or fail to effectively align features at the same spatial location across modalities. 

Based on the above analysis, we redesign the routing path of sparse queries in the sparse BEV alignment methods and propose a novel radar camera alignment model named RCAlign for 3D object detection. The RCAlign not only enables features interaction between modalities during alignment but also guarantees the alignment of features representing the same object across modalities, ensuring consistency and accuracy in feature representation. Specifically, as shown in Fig.~\ref{alignment} (c), we develop a Dual-Route Alignment (DRA) module that updates the sparse queries through two routing paths respectively. Later, the updated sparse queries from two paths are aligned using contrastive loss. Besides, considering the sparsity of the radar BEV features, a Radar Feature Enhancement (RFE) module is proposed, which enhances the representation of radar BEV features by knowledge distillation loss and allows a more effective fusion with image features. The experimental results on the nuScenes benchmark demonstrate the effectiveness of RCAlign. Compared with the existing works, our work has the following contributions.
\begin{itemize}
    \item We propose a novel radar camera alignment model called RCAlign based on the sparse BEV alignment methods for 3D object detection.

    \item In our model, a Dual-Route Alignment module is proposed to  more efficiently align the features of two modalities and perform inter-modal interactions during the alignment process.
                
    \item We design a Radar Feature Enhancement (RFE) module to obtain the enhanced radar BEV features by knowledge distillation loss for better fusion with image features.

    \item Extensive experiments demonstrate that the RCAlign achieves a new state-of-the-art performance on the nuScenes benchmark, with a result of 67.3\% NDS on the nuScenes test set.
\end{itemize}

The remainder of this paper is arranged as follows. In Section \ref{Related work}, we briefly review the background related to our work. Section \ref{Proposed Method} introduces the formulation of our RCAlign. Section \ref{Experiments} gives the experimental results. Finally, the conclusions of the study are drawn in Section \ref{Conclusion}.

\section{Related work} \label{Related work}

In this section, we briefly introduce recent developments related to our study, focusing on three topics: camera-based 3D object detection, radar-camera 3D object detection, and contrastive learning.

\subsection{Camera-based 3D Object Detection} \label{Camera-based 3D Object Detection}
The central focus of recent works on 3D object detection has shifted from PV-based methods \cite{wang2021fcos3d, zhou2019objects, wang2021progressive, kim2023stereoscopic, huang2023obmo} to BEV-based methods \cite{huang2021bevdet, wang2023exploring, hou2024query, wang2022detr3d, li2022bevformer, li2023bevdepth} due to various challenges encountered in PV, such as occlusion, viewpoint transformation and so on. The BEV-based methods can be divided into dense BEV-based methods \cite{li2024bevnext, li2022bevformer, yang2023bevformer, li2023bevdepth, xie2022m, li2023bevstereo, shi2024cobev}, and query-based methods \cite{liu2022petr, wang2023exploring, hou2024query, wang2022detr3d, lin2022sparse4d}. The dense BEV-based methods usually employ the Lift-Splat-Shoot (LSS)\cite{philion2020lift} to transform image features into BEV space. The BEVDet series \cite{huang2021bevdet, huang2022bevdet4d, huang2022bevpoolv2} successfully introduces LSS to 3D object detection and achieves excellent performance. Considering the inaccuracy of LSS for depth estimation, BEVDepth \cite{li2023bevdepth} introduces lidar to supervise the predicted depth. After that, in order to get more accurate information about speeds, the temporal information \cite{huang2022bevdet4d, li2022bevformer, li2023bevdepth} is introduced and further improves the performance of 3D object detection. The query-based methods usually predefine some sparse queries to model the objects and then use the queries to index the related image features. After the introduction of sparse queries for 3D object detection by DETR3D \cite{wang2022detr3d} firstly, numerous methods based on sparse queries have been proposed. PETR \cite{liu2022petr} uses 3D positional encoding to transform image features into 3D position-aware features for end-to-end 3D object detection. In StreamPETR \cite{wang2023exploring}, a memory queue which also consists of sparse queries is designed to store temporal information. Considering that query-based methods do not need to model the background and are more efficient compared to dense BEV-based methods. In the work, we design the RCAlign based on the query-based methods.

\subsection{Radar-camera 3D object detection} \label{Radar-camera 3D object detection}
The radar excels in acquiring both the Doppler velocity and distance of an object, remains virtually unaffected by challenging environmental conditions and is cost-effective. Therefore, radar can serve as an excellent auxiliary sensor to the camera for low-cost 3D object detection. Recently, 3D object detection algorithms based on radar and camera fusion \cite{kim2023crn, lin2024rcbevdet, nabati2021centerfusion, long2023radiant, kim2023craft, wu2023mvfusion} have demonstrated to be effective and have achieved remarkable performance. Similar to the research direction in 3D object detection for multi-view images, researchers initially focused on studying PV space\cite{nabati2021centerfusion, long2023radiant, kim2023craft}. They establish methods for associating radar points with objects in images. The CenterFusion \cite{nabati2021centerfusion} uses the frustum to associate the radar points with the object. Considering the inaccuracy of the radar projection position, RADIANT \cite{long2023radiant} corrects the radar point projection position and designs specific rules to associate radar points with objects in the image using Euclidean coordinates, while CRAFT \cite{kim2023craft} uses polar coordinates for this association. For the BEV-based methods, in order to obtain more accurate image BEV features, the CRN \cite{kim2023crn} utilizes radar assistance to transform image features from PV into BEV. Afterwards they concat the radar features and image features to serve as sparse queries and employ deformable attention to align features from both modalities. Different from the CRN of obtaining sparse queries, the SparseFusion3D \cite{yu2023sparsefusion3d} uses pre-defined sparse queries as an intermediate state to align the features from the two modalities. This type of method does not perform inter-modal interactions during the alignment process. RCBEVDet \cite{lin2024rcbevdet} designs a new backbone to extract radar features, and then uses radar features or image features as sparse queries to query features of another modality to align the features. However, they fail to achieve true alignment between the features from different modalities. In order to alleviate the above problems, we design a dual-route alignment module with contrastive loss which effectively aligns features representing the same object across modalities.

\begin{figure*}[t]
  \centering
  \includegraphics[width=0.95\textwidth]{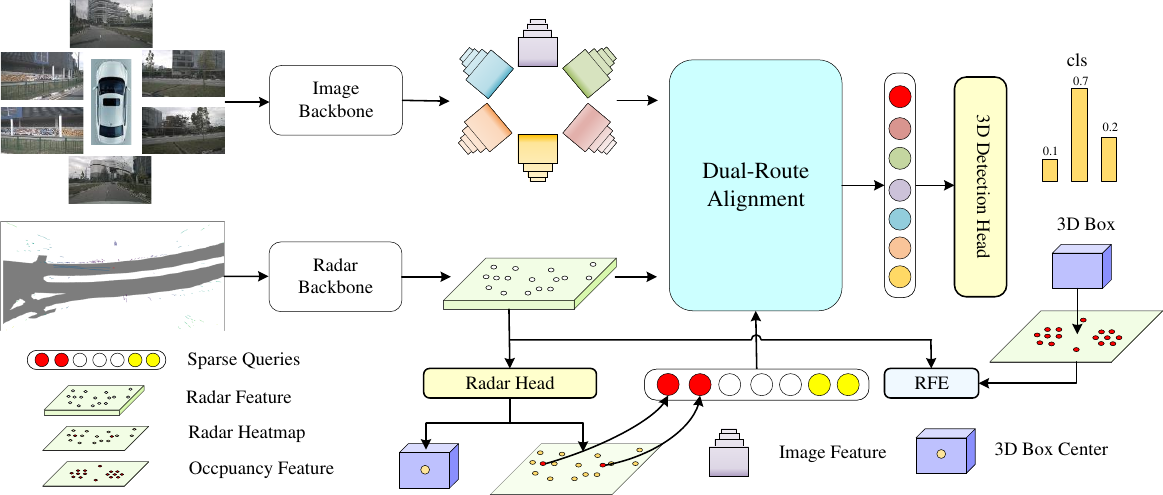}
  \caption{The overall architecture of RCAlign. Multi-view images and radar points are fed into the backbone to extract modal-specific features. Then radar head served as an auxiliary task for predicting both the centre of the 3D boxes and the radar heatmaps. After that, the designed dual-route alignment utilizes sparse queries to align and fuse the features of the two modalities. Finally, the occupancy features obtained by projecting the centre of the 3D boxes onto the BEV grid, along with the radar features, are input into the RFE module to enhance the radar BEV features. Sparse queries are composed of radar queries (red), initial queries (white) and temporal queries (yellow).}
  \label{structure}
\end{figure*}

\subsection{Contrastive Learning} \label{Contrastive Learning}
The contrastive learning \cite{hadsell2006dimensionality, he2020momentum, chen2020simple} aims to pull samples of the same class closer while pushing samples of different classes further apart. It has been applied to the field of object detection to pull in features that represent the same object. SoCo \cite{wei2021aligning} adopts contrastive learning to maximise the similarity of features representing the same object acquired through different augmentation methods. In order to obtain contrastive-aware proposal embeddings, FSCE \cite{sun2021fsce} designs contrastive branch using contrastive learning to maximize the within-category agreement and cross-category disagreement. CAT-Det \cite{zhang2022cat} utilises contrastive learning for data enhancement at the point and object levels, significantly improving detection accuracy. Thus, using contrastive learning enables the similarity of features that represent the same object. Here, we utilize the comparative learning to align the updated sparse queries following the two routing paths.

\section{Proposed Method} \label{Proposed Method}


As analyzed in Sec.~\ref{Introduction}, existing alignment methods for radar and camera fusion either neglect inter-modal features interaction during alignment or fail to align features at corresponding positions. Therefore, we propose a noval Radar Camera Alignment by Contrastive Learning (RCAlign) to alleviate the above problems. In this section, we first describe the overall structure of RCAlign. After that the multimodal feature extraction and the proposed DRA and RFE modules in RCAlign are presented in detail, respectively.

\subsection{Overall Architecture} \label{Overall Architecture}
As shown in Fig.~\ref{structure}, the proposed RCAlign includes an image backbone, a radar backbone, a radar head, a dual-route alignment module, a radar feature enhancement module and a 3D detection head. During the multimodal feature extraction stage, the image backbone and the radar backbone are used to extract multi-view image PV features and radar BEV features respectively. Subsequently, the radar BEV features are passed through the radar head to get the centre of 3D boxes and radar heatmaps. Then, the radar points corresponding to the top $k$ probability values obtained from the radar heatmaps are taken out as part of the sparse queries. In addition to radar queries, sparse queries also contain initial queries and temporal queries. The sparse queries, image PV features and radar BEV features are input into the proposed DRA module to align and fuse the features of the two modalities. After that, the fused features are used to predict the categories and 3D boxes. Finally, considering the sparsity of the radar BEV features, we project the centre of the predicted 3D boxes into the BEV grid as the occupancy features, the occupancy features and the radar BEV features are fed into the RFE module to get the enhanced radar BEV features. The enhanced radar features are used to distill the original radar features, guiding the network to optimize for acquiring denser radar features.

\begin{figure*}[t]
  \centering
  \includegraphics[width=0.9\textwidth]{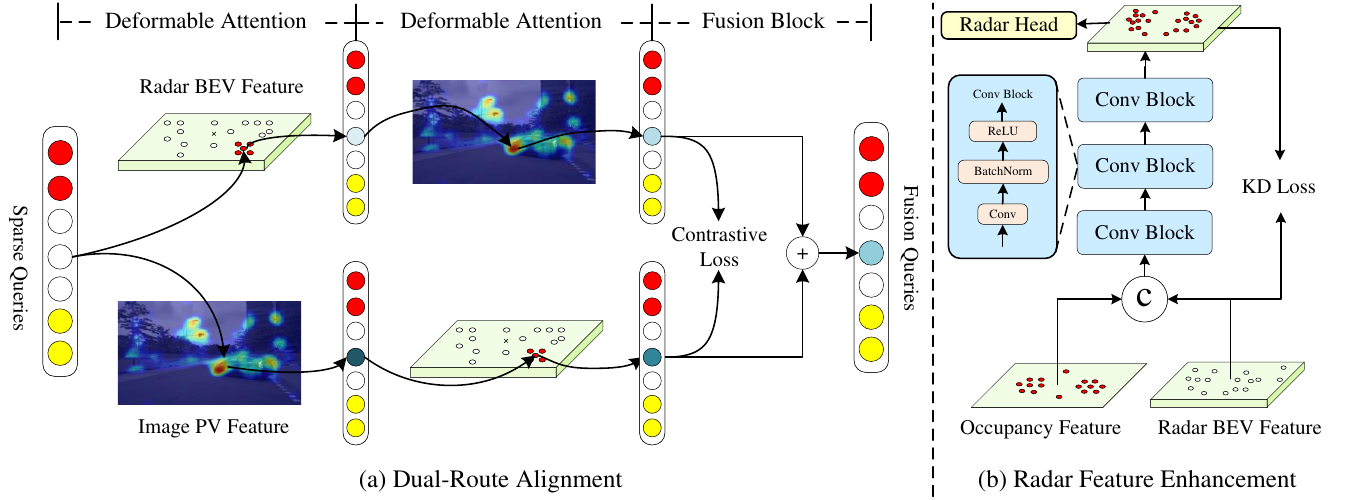}
  \caption{The proposed Dual-Route Alignment (DRA) module (a) and Radar Feature Enhancement (RFE) module (b). The DRA firstly utilizes sparse queries to successively aggregate radar BEV features and image PV features through two separate paths. Then, The updated queries of the two paths are aligned by contrastive loss. Finally, the fusion queries are obtained by element-wise addition of the updated sparse queries from the two paths. For RFE, the occupancy features and the radar BEV features are concatenated and passed through a three-layer conv block to obtain dense radar features. Subsequently, knowledge distillation loss is employed to enhance the original radar features. The KD loss denotes Knowledge Distillation loss.}
  \label{propose}
\end{figure*}

\subsection{Multimodal Feature Extraction}
For image branch, given the images $I \in R^{N \times C \times H \times W}$ from $N$ views, we use the backbone and the FPN \cite{lin2017feature} to extract multi-view image PV features. For radar branch, given the radar data corresponding to the image, we first aggregate five consecutive frames of radar data. During the aggregation process, we introduce velocity to compensate for the radar points position coordinates in the previous frames. Specifically for the radar point at the $t$-1 frames, we compensate the position using the following equation:
\begin{equation}
    \label{position_compensate}
    \begin{aligned}    
    p_{xy}^{t} = p_{xy}^{t-1}+v_{xy}^{t-1}*\Delta{t},
    \end{aligned}
\end{equation}
where $p_{xy}^{t}$ and $p_{xy}^{t-1}$ denote $x,y$ coordinates of the radar point at time $t$ and $t-1$, $v_{xy}^{t-1}$ and $\Delta {t}$ denote the velocity of the radar point at moment $t-1$ and the time interval between $t$ and $t-1$, respectively. Finally, we use PointPillar \cite{lang2019pointpillars} to extract radar BEV features.

\subsection{Dual-Route Alignment} \label{Dual-Route Alignment}
In order to better align the features of the two modalities and to perform the inter-modal
features interaction during the alignment process, we propose the Dual-Route Alignment (DRA) moudle. The DRA achieves alignment and fusion by aggregating image PV features, radar BEV features to the sparse queries. As shown in Fig.~\ref{propose} (a), DRA includes two routing paths, each undergoing two deformable attention \cite{zhu2020deformable} for inter-modal features interaction. Subsequently, the updated sparse queries from the two paths are aligned and fused using the fusion block. Specifically, for the above path, given the sparse queries $z_q$, the radar BEV features $f_r$ and the reference point $p_q$. The process of aggregate radar BEV features can be formulated as:
\begin{equation}
    \label{deform_atten_1}
    \begin{aligned}
      DeformAttn(z_q,p_q,f_r)=\sum_{m=1}^MW_m[\sum_{k=1}^KA_{mqk} \cdot \\ W_m^{\dagger}f_r(p_q+\Delta p_{mqk})],
    \end{aligned}
\end{equation}
where $M$ and $K$ indicate the number of attention heads and sample points respectively. $A_{mqk}$ and $\Delta p_{mqk}$ stand for attention weight and sampling offset respectively. $W_m$ and $W_m^{\dagger}$ denote the weights of the linear. We denote the updated sparse queries as $z_r$. For the image PV features $f_i$, the aggregation of $f_i$ using $z_r$ can be represented as follows:

\begin{equation}
    \label{deform_atten_2}
    \begin{aligned}
      DeformAttn(z_r,p_q,f_i)=\sum_{m=1}^MW_m^{\prime}[\sum_{k=1}^KA_{mqk}^{\prime} \cdot \\ W_m^{\dagger \prime}f_i(p_q+\Delta p_{mqk}^{\prime})],
    \end{aligned}
\end{equation}
where $W_m^{\prime}$ and $W_m^{\dagger \prime}$ indicate the weights of the linear, $A_{mqk}^{\prime}$ and $\Delta p_{mqk}^{\prime}$ indicate attention weight and sampling offset respectively. The updated sparse queries are denoted as $z_{ri}$. For the below path, the same operation is performed, but the image PV features are aggregated before the radar BEV features. The updated sparse queries are denoted as $z_{ir}$. After the aforementioned operation, $z_{ri}$ and $z_{ir}$ are intended to represent the same object at corresponding positions, and their features should have similar semantics. Therefore, we introduce the contrastive learning \cite{radford2021learning} to align the features of the corresponding positions of $z_{ri}$ and $z_{ir}$, which can be calculated as:
\begin{equation}
    \label{contrastive learning}
    \begin{aligned}
       &\widehat{z}_{ri}= \frac{z_{ri}}{||z_{ri}||},  \widehat{z}_{ir}=\frac{z_{ir}}{||z_{ir}||}, \\
       &\mathcal{L}_{CL}= \frac{\mathcal{L}_{CE}(\widehat{z}_{ri} \otimes \widehat{z}_{ir}^t \cdot \tau, I) + \mathcal{L}_{CE}(\widehat{z}_{ir} \otimes \widehat{z}_{ri}^t \cdot \tau, I)}{2},
    \end{aligned}
\end{equation}
where $\widehat{z}_{ri}$ and $\widehat{z}_{ir}$ stand for the normalized features. $\otimes$, $\tau$, and $I$ denote matrix multiplication, logit scale, and target matrix respectively. $\mathcal{L}_{CE}$ represents the cross-entropy loss. Finally, we use element-wise addition to obtain fusion queries $z_f$, which are utilized for 3D object detection task. 

\subsection{Radar Feature Enhancement} \label{Radar Feature Enhancement}
The radar data is obtained by aggregating five consecutive frames, but they are still sparse compared to the lidar. Considering that the number of fused queries exceeds the count of ground truth boxes by a considerable margin, there will be cases where multiple queries predict the same box. Due to most of the predicted boxes representing the same object having low classification probabilities, they have some bias relative to the ground truth boxes but also surround them. Therefore, it is feasible to utilize the predicted centre of the 3D boxes for enhancing radar features. Based on this, we designed the radar feature enhancement module.

As illustrated in Fig.~\ref{propose} (b), the original radar BEV features and occupancy features are fed into RFE to obtain the enhanced radar features. Specifically, we first transform the centre of 3D boxes predicted by the fusion queries $z_f$ into BEV grids to get the occupancy features. Later, the radar BEV features and occupancy features are concatenated on the channel dimension, and the concatenated features are fed into a 3-layer conv block to obtain the enhanced radar BEV features. The conv block is composed of a convolutional layer, followed by batch normalization \cite{ioffe2015batch}, and ReLU \cite{nair2010rectified}. After obtaining the enhanced radar BEV features, on the one hand, they are input to the radar head as the second auxiliary task, sharing parameters with the radar head mentioned in Sec.~\ref{Overall Architecture}. On the other hand, they are used to distil the original radar features by knowledge distillation \cite{hinton2015distilling}, guiding the network to optimize towards the direction where enhanced radar features can be acquired. The knowledge distillation process can be calculated as follows:
\begin{equation}
    \label{knowledge distillation}
    \begin{aligned}
       \mathcal{L}_{KD}=\sum_{k=1}^C\sum_{j=1}^H\sum_{i=1}^W(f_{er}^{ijk}-f_r^{ijk}), \\
    \end{aligned}
\end{equation}
where $H$, $W$ and $C$ represent the height, width, and the number of channels of the radar BEV features, respectively. $f_{er}$ denotes enhanced radar BEV features.

\subsection{Loss Function}
Here, we conclude the loss function including the 3D object detection task losses $\mathcal{L}_{task}$, the losses corresponding to the two radar heads $\mathcal{L}_{RH1}$ and $\mathcal{L}_{RH2}$, the contrastive loss $\mathcal{L}_{CL}$ and the knowledge distillation loss $\mathcal{L}_{KD}$. The final optimization objective can be expressed as:
\begin{equation}
    \label{Loss Function}
    \begin{aligned}    \mathcal{L}=\mathcal{L}_{task}+\lambda_1\mathcal{L}_{RH1}+\lambda_2\mathcal{L}_{RH2}+\lambda_3\mathcal{L}_{CL}+\lambda_4\mathcal{L}_{KD}, 
    \end{aligned}
\end{equation}
where $\lambda_1, \lambda_2, \lambda_3$ and $\lambda_4$ are hyper-parameters.

\begin{table*}[t]
    \centering
    \tabcolsep=2pt
        \caption{Comparison on the nuScenes val set. 'C' and 'R' represent camera and radar respectively. ${\ast}$The backbone benefits from perspective pre-training. }
        \label{val_result}
        \scalebox{1.20}{\begin{tabular}{ c|c c c |c c|c c c c c |c }
        \hline
        Method &Modality &Backbone &Input Size  &NDS $\uparrow$ &mAP $\uparrow$ &mATE $\downarrow$ &mASE $\downarrow$ &mAOE $\downarrow$ &mAVE $\downarrow$ &mAAE $\downarrow$ &FPS $\uparrow$  \\   
	  \hline
        BEVDet \cite{huang2021bevdet} &C &R50 &256 $\times$ 704  &0.392 &0.312 &0.691 &0.272 &0.523 &0.909 &0.247 &15.6 \\
        BEVDepth \cite{li2023bevdepth} &C &R50 &256 $\times$ 704  &0.475 &0.351 &0.639 &0.267 &0.479 &0.428 &0.198 &11.6 \\
        SOLOFusion \cite{park2022time} &C &R50 &256 $\times$ 704  &0.534 &0.427 &0.567 &0.274 &0.411 &0.252 &0.188 &11.4 \\
        StreamPETR \cite{wang2023exploring} &C &R50 &256 $\times$ 704  &0.540 &0.432 &0.581 &0.272 &0.413 &0.295 &0.195 &\textbf{27.1} \\
        SparseBEV \cite{liu2023sparsebev} &C &R50 &256 $\times$ 704  &0.545 &0.432 &0.606 &0.274 &\textbf{0.387} &0.251 &0.186 &- \\
        BEVNeXt \cite{li2024bevnext} &C &R50 &256 $\times$ 704  &0.548 &0.437 &0.550 &0.265 &0.427 &0.260 &0.208 &- \\
        CenterFusion \cite{nabati2021centerfusion} &C+R &DLA34 &448 $\times$ 800  &0.453 &0.332 &0.649 &0.263 &0.535 &0.540 &\textbf{0.142} &- \\
        RCBEV4d \cite{zhou2023bridging} &C+R &Swin-T &256 $\times$ 704  &0.497 &0.381 &0.526 &0.272 &0.445 &0.465 &0.185 &7.5 \\
        CRAFT \cite{kim2023craft} &C+R &DLA34 &448 $\times$ 800  &0.517 &0.411 &0.494 &0.276 &0.454 &0.486 &0.176 &4.1 \\
        CRN \cite{kim2023crn} &C+R &R50 &256 $\times$ 704  &0.560 &0.490 &0.487 &0.277 &0.542 &0.344 &0.197 &20.4 \\
        RCBEVDet \cite{lin2024rcbevdet} &C+R &R50 &256 $\times$ 704  &0.568 &0.453 &0.486 &0.285 &0.404 &0.220 &0.192 &21.3 \\
        \rowcolor{gray!10}
        RCAlign  &C+R &R50 &256 $\times$ 704  &\textbf{0.611} &\textbf{0.537} &\textbf{0.463} &\textbf{0.261} &0.462 &\textbf{0.192} &0.194 &14.6 \\
        \hline
        DETR3D$^{\ast}$ \cite{wang2022detr3d} &C &R101 &900 $\times$ 1600  &0.434 &0.349 &0.716 &0.268 &0.379 &0.842 &0.200 &3.7 \\
        PETR$^{\ast}$ \cite{liu2022petr} &C &R101 &900 $\times$ 1600  &0.442 &0.370 &0.711 &0.267 &0.383 &0.865 &0.201 &1.7 \\
        BEVFormer$^{\ast}$  \cite{li2022bevformer} &C &R101 &900 $\times$ 1600  &0.517 &0.416 &0.673 &0.274 &0.372 &0.394 &0.198 &1.7 \\
        BEVDepth \cite{li2023bevdepth} &C &R101 &512 $\times$ 1408  &0.535 &0.412 &0.565 &0.266 &0.358 &0.331 &0.190 &5.0 \\
        SOLOFusion \cite{park2022time} &C &R101 &512 $\times$ 1408  &0.582 &0.483 &0.503 &0.264 &0.381 &0.246 &0.207 &- \\
        SparseBEV$^{\ast}$ \cite{liu2023sparsebev} &C &R101 &512 $\times$ 1408  &0.592 &0.501 &0.562 &0.265 &0.321 &0.243 &0.195 &- \\
        StreamPETR$^{\ast}$ \cite{wang2023exploring} &C &R101 &512 $\times$ 1408  &0.592 &0.504 &0.569 &0.262 &\textbf{0.315} &0.257 &0.199 &6.4 \\
        BEVNeXt$^{\ast}$ \cite{li2024bevnext} &C &R101 &512 $\times$ 1408  &0.597 &0.500 &0.487 &0.260 &0.343 &0.245 &0.197 &4.4 \\
        MVFusion$^{\ast}$ \cite{wu2023mvfusion} &C+R &R101 &900 $\times$ 1600  &0.455 &0.380 &0.675 &0.258 &0.372 &0.833 &0.196 &- \\
        CRN \cite{kim2023crn} &C+R &R101 &512 $\times$ 1408  &0.592 &0.525 &0.460 &0.273 &0.443 &0.352 &\textbf{0.180} &\textbf{7.2} \\
        \rowcolor{gray!10}
        RCAlign$^{\ast}$ &C+R &R101 &512 $\times$ 1408  &\textbf{0.645} &\textbf{0.570} &\textbf{0.457} &\textbf{0.257} &0.319 &\textbf{0.186} &0.187 &5.4 \\
        \hline
        \end{tabular}}
\end{table*}

\begin{table*}[thbp]
    \centering
    \caption{Comparison of Per-class AP on nuScenes val set. 'C.V.', 'Ped.', 'M.C.', and 'T.C.' denote construction vehicle, pedestrian, motorcycle, and traffic cone, respectively.}
    \label{Per_Class}
    \scalebox{1.22}{\begin{tabular}{c| c c c c c c c c c c|c}
          \hline
           Method  &Car &Truck &Bus &Trailer & C.V. &Ped. &M.C. &Bicycle & T.C. & Barrier &mAP \\
          \hline
          CenterFusion \cite{nabati2021centerfusion}   &0.524 &0.265 &0.362 &0.154 &0.055 &0.389 &0.305 &0.299 &0.563 &0.470 &0.332 \\
          RCBEV4d \cite{lin2024rcbevdet}  &0.683 &0.323 &0.369 &0.148 &0.108 &0.443 &0.357 &0.270 &0.552 &0.557 &0.381 \\
          CRAFT \cite{kim2023craft}  &0.696 &0.376 &0.473 &0.201 &0.107 &0.462 &0.395 &0.310 &0.571 &0.511 &0.411 \\
          CRN \cite{kim2023crn}  &0.736 &0.445  &0.556  &0.220  &0.154  &0.502  &0.547 &0.489  &0.614  &\textbf{0.638}  &0.490  \\
          \rowcolor{gray!10}
          RCAlign  & \textbf{0.777} &\textbf{0.488} &\textbf{0.589} &\textbf{0.234} &\textbf{0.231} &\textbf{0.617} &\textbf{0.596} &\textbf{0.534} &\textbf{0.667} &0.636 &\textbf{0.537} \\ 
          \hline
    \end{tabular}}
\end{table*}

\section{Experiments} \label{Experiments}

In this section, we conduct extensive experiments on our RCAlign using the nuScenes dataset\cite{caesar2020nuscenes}. First, we compare RCAlign with other other relevant camera-based and radar camera fusion methods in 3D object detection and tracking tasks. After that, rigorous ablation experiments are given to demonstrate the effectiveness of the proposed module. Then, we perform key parametric analyses that may affect the performance of model. Finally, we conduct a robustness analysis experiment. The above experimental results validate the effectiveness of the proposed model.

\subsection{Experimental Settings} \label{Experimental_Settings}

\textbf{Datasets and Metrics.} The nuScenes dataset \cite{caesar2020nuscenes} is a publicly available large-scale dataset for autonomous driving, comprising 1000 driving scenes. Among these, 750 scenes are designated for training, 150 for validation, and the remaining 150 for testing. Each scene in the nuScenes dataset spans a duration of 20 seconds and is annotated using a frequency of 2 Hz. The data in each scene is captured using 6 cameras that provide a full 360-degree field of view, along with 5 radars and 1 lidar. The annotated 3D boxes are categorized into 23 classes initially and aggregated into 10 categories for the 3D object detection task during evaluation. The mean Average Precision (mAP), nuScenes Detection Score (NDS), mean Average Translation Error (mATE), mean Average Scale Error (mASE), mean Average Orientation Error (mAOE), mean Average Velocity Error (mAVE) and mean Average Attribute Error (mAAE) are used as the evaluation metrics for 3D object detection. We compare the Average Multi-Object Tracking Accuracy (AMOTA), Average Multi-Object Tracking Precision (AMOTP), number of False Positives (FP), number of False Negatives (FN) and number of Identity Switches (IDS) with those of other methods for 3D object tracking. 

\textbf{Implementation Details.} We use ResNet50 \cite{he2016deep}, ResNet101 and V2-99 \cite{lee2019energy} backbones to extract image features. The R50 and R101 are used for nuScenes val set, which are pre-train on ImageNet \cite{deng2009imagenet} and nuImages \cite{caesar2020nuscenes}. The V2-99 is used for nuScenes test set and initialized from DD3D \cite{park2021pseudo}. We use the [$x$, $y$, $z$, $vx_{comp}$, $vy_{comp}$, $timestamp$] as the radar features. The point cloud range is set to [-51.2m, 51.2m] for X and Y axis, and [-5.0m, 3.0m] for Z axis. The radar backbone follows the architecture of FUTR3D \cite{chen2023futr3d}. The radar head adopts CenterNet \cite{zhou2019objects}, and as an auxiliary task, we simplify it by retaining only the prediction of the heatmap and the centers of 3D boxes. The 3D detection head is the same as DETR3D \cite{wang2022detr3d}. 

During training, 2D supervised losses \cite{wang2023focal} and query denoies \cite{li2022dn} are introduced. All experiments were conducted without the use of CBGS \cite{zhu2019class} and TTA strategies. For image and BEV data augmentation, we follow the PETR \cite{liu2022petr}, the radar data corresponds to the transformation when performing BEV data augmentation. Following StreamPETR \cite{wang2023exploring} when compared with other state-of-the-art methods, the model is trained for 60 epochs. During ablation experiments, it is trained for 24 epochs. AdamW \cite{loshchilov2017decoupled} optimizer is used to optimize the model with the learning rate set to 4e-4. The learning rate is updated by the cosine annealing strategy. The models are trained end-to-end. The number of radar queries is set to 30. The $\lambda_1$, $\lambda_2$ and $\lambda_3$ are set to 1, the $\lambda_4$ is set to 5. We implement RCAlign through the mmdetection3D framework and train it on NVIDIA 3090 GPUs. 

\subsection{Comparison with State-of-the-Art Methods} \label{Comparison with State-of-the-Art Methods}


In this section, we first present the results of RCAlign and other state-of-the-art methods for 3D object detection on the nuScenes val set and test set. Following that, visualisation results of RCAlign on the nuScenes val set are presented. Finally, we give the results of RCAlign in 3D object tracking.

\textbf{3D Object Detection on val set.} We compare the RCAlign with the state-of-the-art methods on nuScenes val for 3D object detection in Tab.~\ref{val_result}. \textbf{(1)} With the R50 \cite{he2016deep} backbone and the input resolution of 256$\times$704, RCAlign outperforms both existing camera-based methods and radar-camera fusion methods, including the latest state-of-the-art methods CRN \cite{kim2023crn} and RCBEVDet \cite{lin2024rcbevdet}. The NDS and mAP are significant improvements compared to RCBEVDet, with increases of 4.3\% and 8.4\%, respectively. \textbf{(2)} To the best of our knowledge, this may be the first radar-camera fusion algorithm to achieve an NDS exceeding 60\% in real-time 3D object detection. \textbf{(3)} Furthermore, RCAlign with R50 backbone can even outperform CRN using the R101 backbone with the input resolution of 512$\times$1408. \textbf{(4)} When the model size and image size increase to R101 and 512$\times$1408, RCAlign still can outperform all previous camera-based or radar and camera fusion methods. Alongside the significant improvements in NDS and mAP, mAVE has also made remarkable strides. The improvement of mAVE can be attributed to the ability of radar to capture information about the speed of objects. 

\textbf{Pre-class AP on val set.} In order to better demonstrate the performance of RCAlign, we analyse the pre-class AP for the radar and camera fusion models in Tab.~\ref{Per_Class}. From the table we can see that all the classes except barriers have been improved. Especially for pedestrians, which always is small or clustered, there has been significant improvement, with increasing by 11.5\% compared to CRN. In the case of barriers, comparison with CRN shows almost equal performance. In terms of mAP, we have also seen a significant improvement compared to other radar and camera fusion methods.

\begin{table*}[h]
    \centering
    \tabcolsep=2pt
        \caption{Comparison on the nuScenes test set. 'C'and 'R' represent camera and radar respectively.}
        \label{test_result}
        \scalebox{1.25}{\begin{tabular}{ c|c c c|c c|c c c c c}
        \hline
        Method &Modality &Backbone &Input Size  &NDS $\uparrow$ &mAP $\uparrow$ &mATE $\downarrow$ &mASE $\downarrow$ &mAOE $\downarrow$ &mAVE $\downarrow$ &mAAE $\downarrow$  \\   
	  \hline 
        DETR3D \cite{wang2022detr3d} &C &V2-99 &900 $\times$ 1600  &0.479 &0.412 &0.641 &0.255 &0.394 &0.845 &0.133  \\
        BEVFormer \cite{li2022bevformer} &C &V2-99 &900 $\times$ 1600  &0.569 &0.481 &0.582 &0.256 &0.375 &0.378 &0.126  \\
        PETRv2 \cite{liu2023petrv2} &C &V2-99 &640 $\times$ 1600 &0.582 &0.490 &0.561 &0.243 &0.361 &0.343 &0.120  \\
        BEVDepth \cite{li2023bevdepth} &C &V2-99 &640 $\times$ 1600  &0.600 &0.503 &0.445 &0.245 &0.378 &0.320 &0.126  \\
        SOLOFusion \cite{park2022time} &C &ConvNeXt-B &640 $\times$ 1600  &0.619 &0.540 &0.453 &0.257 &0.376 &0.276 &0.148  \\
        StreamPETR \cite{wang2023exploring} &C &V2-99 &640 $\times$ 1600 &0.636 &0.550 &0.479 &0.239 &\textbf{0.317} &0.241 &0.119  \\
        SparseBEV \cite{liu2023sparsebev} &C &V2-99 &640 $\times$ 1600  &0.636 &0.556 &0.485 &0.244 &0.332 &0.246 &0.117  \\
        BEVNeXt \cite{li2024bevnext} &C &V2-99 &640 $\times$ 1600  &0.642 &0.557 &0.409 &0.241 &0.352 &0.233 &0.129  \\
        CenterFusion \cite{nabati2021centerfusion} &C+R &DLA34 &448 $\times$ 800 &0.449 &0.326 &0.631 &0.261 &0.516 &0.614 &0.115  \\
        RCBEV \cite{zhou2023bridging} &C+R &Swin-T &256 $\times$ 704  &0.486 &0.406 &0.484 &0.257 &0.587 &0.702 &0.140  \\
        MVFusion \cite{wu2023mvfusion} &C+R &V2-99 &640 $\times$ 1600  &0.517 &0.453 &0.569 &0.246 &0.379 &0.781 &0.128  \\
        CRAFT \cite{kim2023craft} &C+R &DLA34 &448 $\times$ 800  &0.523 &0.411 &0.467 &0.268 &0.456 &0.519 &0.114  \\
        CRN \cite{kim2023crn} &C+R &ConvNeXt-B &640 $\times$ 1600  &0.624 &0.575 &0.416 &0.264 &0.456 &0.365 &0.130  \\
        RCBEVDet \cite{lin2024rcbevdet} &C+R &V2-99 &640 $\times$ 1600  &0.639 &0.550 &0.390 &\textbf{0.234} &0.362 &0.259 &\textbf{0.113}  \\
        \rowcolor{gray!10}
        RCAlign &C+R &V2-99 &640 $\times$ 1600  &\textbf{0.673} &\textbf{0.606} &\textbf{0.385} &0.241 &0.360 &\textbf{0.191} &0.123  \\
        \hline
        \end{tabular}}
\end{table*}

\begin{figure*}[h]
  \centering
  \includegraphics[width=1\textwidth]{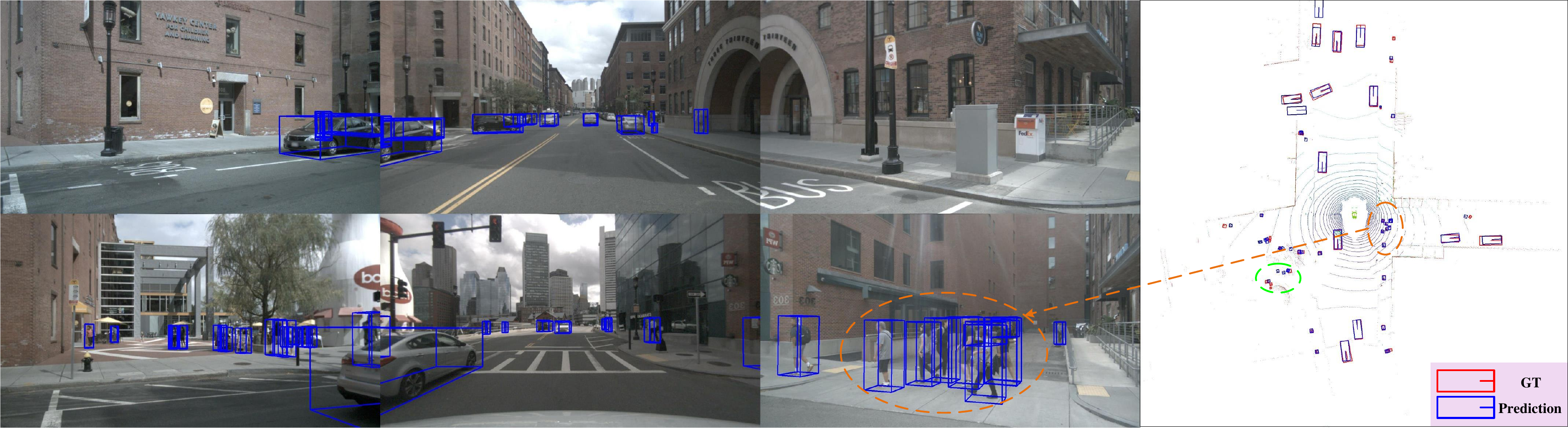}
  \caption{Visualisation results of RCAlign. The red and blue boxes indicate ground truth and prediction, respectively. The orange and green circles indicate examples of predicted successes and failures in dense populations, respectively. The GT denotes ground truth.}
  \label{vis_result}
\end{figure*}

\begin{table*}[t]
    \centering
        \caption{Comparison of 3D object tracking on nuScenes test set.}
        \label{3D_object_tracking}
        \scalebox{1.22}{\begin{tabular}{ c|c|c| c c|c c c}
          \hline
          Methods &Modality &Backbone & AMOTA $\uparrow$ & AMOTP $\downarrow$ & FP $\downarrow$ & FN  $\downarrow$ & IDS $\downarrow$ \\
          \hline
          UVTR \cite{li2022unifying} & C &V2-99 & 51.9 & 1.125 & 14994 &39209 &2204  \\
          StreamPETR \cite{wang2023exploring}  & C & ConvNext-B & 56.6 & 0.975 & 21268 & 31484 & 784\\
          BEVNeXt \cite{li2024bevnext} &C &V2-99 & 57.8 & 0.917 & - & - & \textbf{519}\\
          CRN \cite{kim2023crn} &C+R &ConvNeXt-B & 56.9 & \textbf{0.809} & 16822 & 41093 & 946\\
          \rowcolor{gray!10}
          RCAlign  &C+R &V2-99 & \textbf{60.5} & 0.901 & 1\textbf{4153} & \textbf{29266} & 689\\
          \hline
        \end{tabular}}
\end{table*}

\textbf{3D Object Detection on test set.} For the nuScenes test set, we use the V2-99 \cite{lee2019energy} backbone and the input resolution is 640$\times$1600. Table \ref{test_result} demonstrates that RCAlign has a notable advancement over existing state-of-the-art methods. When compared to RCBEVDet, there is an improvement of 3.4\% NDS and 5.6\% mAP. Additionally, RCBEVDet \cite{lin2024rcbevdet} improves baseline (BEVDepth) results by 3.4\% NDS and 3.5\% mAP. In contrast, RCAlign enhances baseline (StreamPETR) results by 3.7\% NDS and 5.6\% mAP, which is surpassing RCBEVDet. The above analyses show that RCAlign can achieve outstanding results, thus validating the effectiveness of the proposed model.

\begin{figure*}[thbp]
  \centering
  \includegraphics[width=0.75\textwidth]{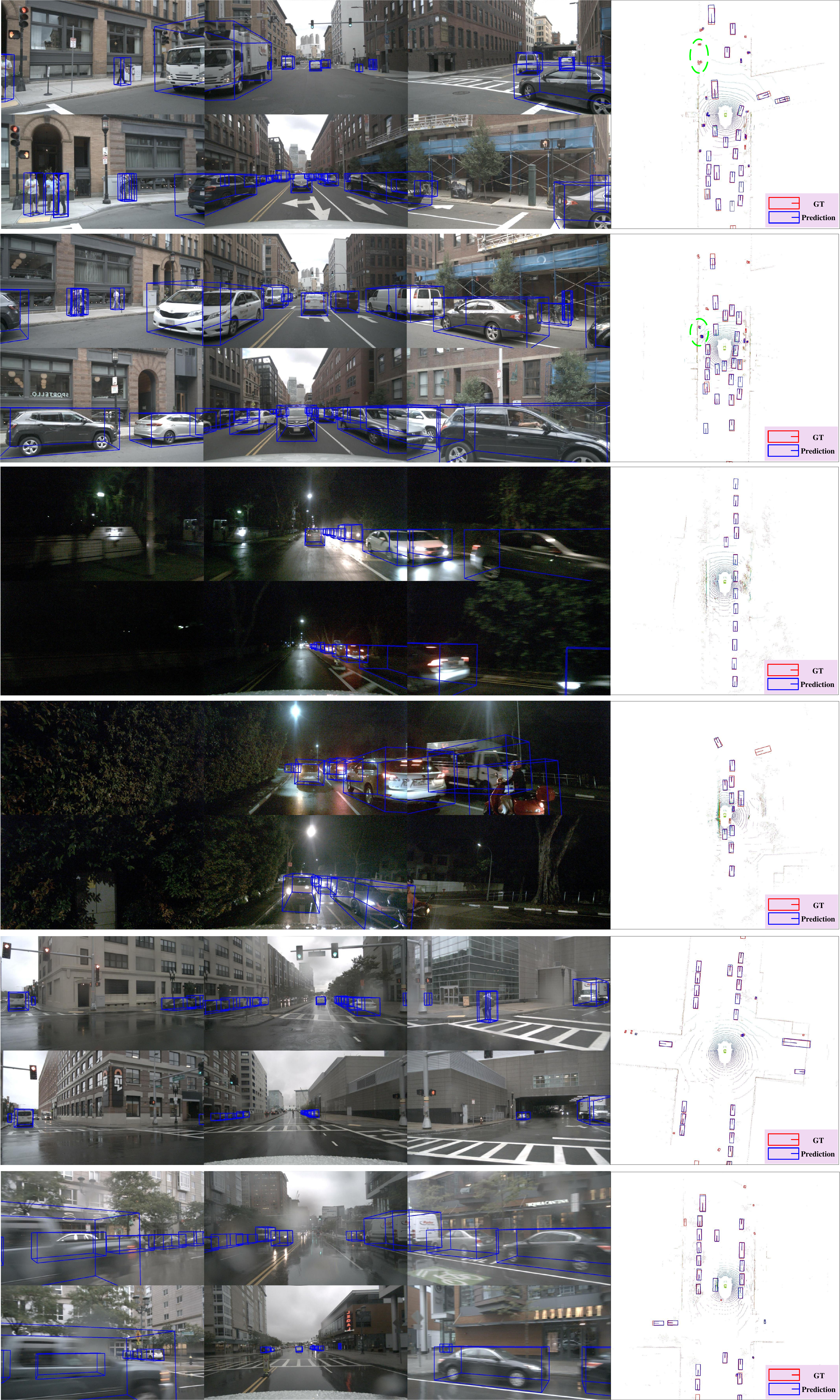}
  \caption{More visualisation results. The first two rows indicate day, the middle two rows indicate night, and the last two rows indicate rain.}
  \label{vis_more}
\end{figure*}

\textbf{Visualisation Results.} We show the visualisation results in the camera view (left) and BEV view (right) in Fig.~\ref{vis_result}. In the BEV view, RCAlign can present excellent detection results, particularly for the densely gathered crowd highlighted in the orange circle. However, as indicated by the green circle, there are some instances of failure, where some objects are either misdetected or missed. This may be due to the distance from the vehicle resulting in poorer detection results. More visualisation results for different weather conditions are provided in Fig.~\ref{vis_more}. From the figure, we can see that RCAlign can  obtain detection results closer to the ground truth in various weather conditions.

\textbf{3D Object Tracking on test set.} As shown in Tab.~\ref{3D_object_tracking}, when compared to the other state-of-the-art methods with the V2-99 backbone, RCAlign achieves the best performance in both AMOTA and AMOTP. The AMOTA can be improved by 2.7\% compared to BEVNext. With the addition of the ConvNext-B backbone, RCAlign still performs best on AMOTA, which is the most dominant evaluation metric in nuScenes test set for 3D object tracking. RCAlign improves by 3.6\% in AMOTA compared to CRN. For other metrics, RCAlign achieves the best or just below the best performance.

\begin{table*}[t]
    \centering
    \caption{Ablation Study of proposed DRA and RFE. C, R, DA, SDA, RH and SRH denote camera branch, radar branch, deformable attention, second deformable attention, radar head and second radar head respectively.}
    \label{ablation_study}
     \scalebox{1.22}{\begin{tabular}{c|c|c|c|c|c|c|c|c|c c |c c}
          \hline
           \multirow{2}{*}{} &\multirow{2}{*}{C} &\multirow{2}{*}{R} &\multicolumn{3}{c|}{DRA} &\multirow{2}{*}{RH} &\multicolumn{2}{c|}{RFE} &\multirow{2}{*}{NDS$\uparrow$} &\multirow{2}{*}{mAP$\uparrow$} &\multirow{2}{*}{mATE$\downarrow$} & \multirow{2}{*}{mAOE $\downarrow$} \\
          \cline{4-6} \cline{8-9}
          &  & &DA &SDA &CL & &SRH &KD & & & & \\
          \hline
          \multirow{5}{*}{(a)} &\CheckmarkBold & & & & & & & &0.518 &0.412 &0.609 &0.516 \\
          & \CheckmarkBold & \CheckmarkBold &\CheckmarkBold & & & & & &0.569 &0.486 &0.525 &0.551 \\
          & \CheckmarkBold & \CheckmarkBold &\CheckmarkBold & &\CheckmarkBold & & & &0.576 &0.496 &0.522 &0.537 \\
          & \CheckmarkBold & \CheckmarkBold &\CheckmarkBold &\CheckmarkBold & & & & &0.583 &0.501 &0.517 &0.503\\
          \rowcolor{gray!10}
          & \CheckmarkBold & \CheckmarkBold &\CheckmarkBold &\CheckmarkBold &\CheckmarkBold & & & &0.584 &0.507 &0.512 &0.515 \\
          \hline \hline
          \multirow{3}{*}{(b)} & \CheckmarkBold & \CheckmarkBold &\CheckmarkBold &\CheckmarkBold &\CheckmarkBold &\CheckmarkBold & & &0.586 &0.504 &0.523 &\textbf{0.476} \\
          & \CheckmarkBold & \CheckmarkBold &\CheckmarkBold &\CheckmarkBold &\CheckmarkBold &\CheckmarkBold &\CheckmarkBold & &0.588 &0.508 &0.519 &0.479 \\
          \rowcolor{gray!10}
          & \CheckmarkBold &\CheckmarkBold &\CheckmarkBold &\CheckmarkBold &\CheckmarkBold &\CheckmarkBold &\CheckmarkBold &\CheckmarkBold &\textbf{0.592} &\textbf{0.515} &\textbf{0.502} &0.487 \\
          \hline
    \end{tabular}}
\end{table*}

\begin{table*}[t]
  \centering
  \begin{minipage}[t]{0.48\textwidth}
    \centering
    \caption{Comparison of using different sampling methods when selecting radar points. SM and FPS denote sampling method and farthest point sampling, respectively.}
    \scalebox{1.22}{\begin{tabular}{ c|c c|c c}
      \hline
      SM & NDS $\uparrow$ & mAP $\uparrow$ & mATE $\downarrow$ & mAOE $\downarrow$ \\
      \hline
      Random & 0.585 & 0.500 & 0.505 & 0.494  \\
      FPS    & 0.588 & 0.512 & 0.518 & 0.515  \\
      \rowcolor{gray!10}
      Topk   & \textbf{0.592} & \textbf{0.515} & \textbf{0.502} & \textbf{0.487}  \\
      \hline
    \end{tabular}}
    \label{sampleing method}
  \end{minipage}
  \hfill
  \begin{minipage}[t]{0.48\textwidth}
    \centering
    \caption{Comparison of using different alignment losses (AL) to align sparse queries from two paths. Cos and CL denote cosine loss and contrastive loss, respectively.}
    \scalebox{1.22}{\begin{tabular}{c|c c|c c}
      \hline
      AL & NDS $\uparrow$ & mAP $\uparrow$ & mATE $\downarrow$ & mAOE $\downarrow$ \\
      \hline
      L1  & 0.514 & 0.421 & 0.594 & 0.693  \\
      Cos & 0.576 & 0.497 & 0.536 & 0.517  \\
      \rowcolor{gray!10}
      CL  & \textbf{0.592} & \textbf{0.515} & \textbf{0.502} & \textbf{0.487}  \\
      \hline
    \end{tabular}}
    \label{alignment loss}
  \end{minipage}
\end{table*}

\subsection{Ablation Study} \label{Ablation Study}

In order to verify the validity of important components of RCAlign, in this part, we give ablation experiments for DRA, REF and their containing sub-modules.

\textbf{Effectiveness of DRA.} In Tab.~\ref{ablation_study} (a), we validate the effectiveness of each sub-module of the DRA. \textbf{(1)} Firstly, we present the results of StreamPETR \cite{wang2023exploring} with deformable attention (baseline) in the first line. Then, a radar branch is added to the baseline (Rbaseline), and the features from both modalities are aggregated using deformable attention, resulting in a significant improvement over baseline by 5.1\% NDS and 7.4\% mAP. This part proves that radar can indeed assist 3D object detection for better performance. \textbf{(2)} Afterwards, the contrastive loss or the second deformable attention is added to Rbaseline, and it is clear from lines 3 and 4 that adding any of the sub-module of DRA can improve the performance of Rbaseline on both NDS and mAP. The same results are observed for mATE and mAOE. The above results demonstrate that the proposed alignment strategy and the inter-modal features interaction are beneficial for Rbaseline. \textbf{(3)} Finally, both sub-modules are added to the Rbaseline, resulting in further performance improvements. \textbf{(4)} In summary, by adding the DRA, the model outperforms the baseline by 6.6\% NDS and 9.5\% mAP and surpasses the Rbaseline by 1.5\% NDS and 2.1\% mAP. The experimental results on nuScenes dataset validate the effectiveness of DRA. 

\textbf{Effectiveness of RH and RFE.} We conducted ablation experiments on the radar head (RH) and RFE module, the results are presented in Tab.~\ref{ablation_study} (b). \textbf{(1)} We first introduce the RH module based on the DRA module, which resulted in a 0.2\% gain in NDS. \textbf{(2)} After that, we use the second radar head (SRH) to classify and regress the enhanced radar BEV features, the NDS and mAP can be further improved, which increased by 0.2\% NDS and 0.4\% mAP. \textbf{(3)} Furthermore, by exploiting knowledge distillation loss, NDS and mAP are improved by 0.4\% and 0.7\%, respectively. After adding the RH and RFE, Both NDS and mAP increased by 0.8\%. Thus, the effectiveness of RH and RFE is verified. \textbf{(4)} It is worth noting that when compared with the baseline, the proposed RCAlign enhances NDS by 7.4\% and mAP by 10.3\%. Even for the designed Rbaseline, RCAlign has achieved significant improvements, with NDS and mAP increasing by 2.3\% and 2.9\%, respectively.

\subsection{Parametric Analysis} \label{Parametric Analysis}

Here, we analyse the important parameters in the model. This includes the Impact of different radar point sampling methods, the impact of different alignment losses, and the impact of contrastive loss and distillation loss weights.

\begin{table*}[h]
  \centering
  \begin{minipage}[t]{0.48\textwidth}
    \centering
    \caption{Effect of constrastive loss wight $\lambda_3$.}
    \scalebox{1.22}{\begin{tabular}{ c|c c|c c}
      \hline
      $\lambda_3$ & NDS $\uparrow$ & mAP $\uparrow$ & mATE $\downarrow$ & mAOE $\downarrow$ \\
      \hline
      0.1  & 0.583 & 0.503 & 0.523 & 0.516  \\
      \rowcolor{gray!10}
      1    & \textbf{0.592} & \textbf{0.515} & \textbf{0.502} & \textbf{0.487}  \\
      10   & 0.585 & 0.507 & 0.531 & 0.494 \\
      \hline
    \end{tabular}}
    \label{cl loss weight}
  \end{minipage}
  \hfill
  \begin{minipage}[t]{0.48\textwidth}
    \centering
    \caption{Effect of distillation loss wight $\lambda_4$.}
    \scalebox{1.22}{\begin{tabular}{c|c c|c c}
      \hline
      $\lambda_4$ & NDS $\uparrow$ & mAP $\uparrow$ & mATE $\downarrow$ & mAOE $\downarrow$ \\
      \hline
      1  &0.581  &0.502  &0.517  &0.528   \\
      \rowcolor{gray!10}
      5  & \textbf{0.592} & \textbf{0.515} & \textbf{0.502} & \textbf{0.487}  \\
      10 &0.580  &0.502  &0.527  &0.523  \\
      \hline
    \end{tabular}}
    \label{KD loss weight}
  \end{minipage}
\end{table*}

\textbf{Impact of sampling methods.} As depicted in Fig.~\ref{structure}, radar points are selected to be part of the sparse queries after the radar head. Here, we analyze the impact of different sampling methods for radar points on the experimental results. As shown in Tab.~\ref{sampleing method}, \textbf{(1)} we start with a random sampling approach, the NDS and mAP are both terrible. \textbf{(2)} After that, we use the farthest point sampling (FPS), which is employed in PointNet++  \cite{qi2017pointnet++}, we take the radar point corresponding to the highest probability value as the initial point. Compared with the random sampling, there are some improvements in NDS and mAP, but some reduction in mATE and mAOE. \textbf{(3)} Finally, we attempt to select radar points corresponding to the top $k$ probability values (Topk) obtained from the radar heatmap. Compared to the previous two methods, this approach led to improvements in NDS, mAP, mATE, and mAOE. \textbf{(4)} Therefore, we choose Topk as the selection method for radar points.

\textbf{Impact of alignment Loss.} We experiment with three different alignment losses to align sparse queries after the two route paths. As shown in Tab.~\ref{alignment loss}, \textbf{(1)} we initially use the L1 loss to align the two sparse queries in DRA, but it produce poorer results. \textbf{(2)} When the alignment loss is changed to cosine loss, there is a significant improvement in overall metrics. This is due to that although the features sampled by the two routing paths are the same, the different order of sampling means that the final feature representations should only have semantic similarity, rather than forcing the feature values to be identical as well. \textbf{(3)} After that, we modify the alignment loss to a contrastive loss, and we can see that all the metrics are further improved. In contrast to the cosine loss, contrastive loss not only pulls the similarity of features at the same location but also simultaneously pushes apart features at different locations to some extent. \textbf{(4)} Based on the above analysis, we choose contrastive loss for the alignment of the two route paths. 

\textbf{Impact of $\lambda$.} We conduct experiments on how the weight of contrastive loss and knowledge distillation loss would affect the performance of the model in the Eq.~\ref{Loss Function}. As shown in Tab.~\ref{cl loss weight}, with the $\lambda_3$ increases, all metrics first increase and then decrease, and the model achieves optimal performance when the weight of the contrastive loss $\lambda_3$ is set to 1. Similarly, Tab.~\ref{KD loss weight} shows that the model performs optimally when the weight of the knowledge distillation loss $\lambda_4$ is set to 5.

\subsection{Robustness Analyse} \label{Robustness Analyse}

\begin{table}[h]
    \centering
        \caption{The robustness analysis of different lighting and weather conditions using mAP metric.}
        \label{robustness}
        \scalebox{1.15}{\begin{tabular}{ c|c | c c c c}
        \hline
         &Modality &Sunny &Rainy  &Day &Night  \\   
	  \hline
        CenterPoint &L &0.629 &0.592  &0.628 &0.354 \\
        \hline
        RCBEV &C+R &0.361 &0.385  &0.371 &0.155 \\
        CRN   &C+R &0.548 &0.570  &0.551 &0.304 \\
        \rowcolor{gray!10}
        RCAlign &C+R &\textbf{0.567} &\textbf{0.610} &\textbf{0.575} &\textbf{0.357} \\
        \hline
        \end{tabular}}
\end{table}

A robust detection algorithm is crucial for enhancing vehicle safety in autonomous driving. Therefore, we conducte robustness experiments under different weather and lighting conditions. Following the CRN \cite{kim2023crn}, we use the R101 backbone and the input resolution of 512$\times$1408, as shown in Tab.~\ref{robustness}, when compared to the radar camera fusion methods, RCAlign achieves state-of-the-art performance across various weather and lighting scenes. Additionally, due to radar is not affected by weather or lighting conditions, RCAlign even outperforms lidar-based method (CenterPoint) in rainy and night scenes. The above experimental results demonstrate that RCAlign is both effective and robustness in different weather conditions.

\section{Conclusion} \label{Conclusion}
In this paper, we propose a novel alignment model RCAlign for radar and camera fusion. Firstly, we designed the DRA, which consists of two deformable attention and a fusion block. The two deformable attention are employed for inter-modal features interaction. The fusion module is utilized to align features using contrastive loss and fuse features from two modalities through element-wise addition. Afterwards, considering the sparsity of radar features, we design a radar feature enhancement module that uses predicted the centre of the 3D boxes to densify the original radar features by knowledge distillation loss. Besides, we introduce the radar head acting on the original radar features and the enhanced radar features as an auxiliary task. Finally, the designed model is optimized by combining the above losses and the 3D object detection task losses. Extensive experiments conducted on the nuScenes benchmark illustrate that the proposed RCAlign currently achieves a new state-of-the-art performance. At the same time, the rigorous ablation experiments demonstrate the effectiveness of our proposed DRA and RFE.


\section*{Acknowledgments}
This research was supported by the National Natural Science Foundation of China under Grant 62272035.

\bibliographystyle{IEEEtran}
\bibliography{egbib}



\end{document}